\documentclass[11pt,a4paper]{article}
\usepackage[hyperref]{acl2021}
\usepackage{times}
\usepackage{latexsym}
\usepackage{booktabs}

\usepackage{booktabs}
\usepackage{graphicx}
\usepackage{multirow}
\aclfinalcopy
\def\aclpaperid{3852}

\title{Don't Rule Out Monolingual Speakers:\\A Method For Crowdsourcing Machine Translation Data}

\author{Rajat Bhatnagar, Ananya Ganesh \and Katharina Kann \\
  University of Colorado Boulder\\
  \texttt{\{rajat.bhatnagar, ananya.ganesh, katharina.kann\}@colorado.edu}}
  
\date{}

\begin{document}
\maketitle
\begin{abstract}
High-performing machine translation (MT) systems can help overcome language barriers while making it possible for everyone to communicate and use language technologies in the language of their choice. However, such systems require large amounts of parallel sentences for training, and translators can be difficult to find and expensive. Here, we present a data collection strategy for MT which, in contrast, is cheap and simple, as it does not require bilingual speakers. Based on the insight that humans pay specific attention to movements, we use graphics interchange formats (GIFs) as a pivot to collect parallel sentences from monolingual annotators. We use our strategy to collect data in Hindi, Tamil and English. As a baseline, we also collect data using images as a pivot. We perform an intrinsic evaluation by manually evaluating a subset of the sentence pairs and an extrinsic evaluation by finetuning mBART \citep{liu2020multilingual} on the collected data. We find that sentences collected via GIFs are indeed of higher quality.
\end{abstract}

\section{Introduction}
Machine translation (MT) -- automatic translation of text from one natural language into another --
 provides access to information written in foreign languages and enables communication between speakers of different languages. However, developing high performing MT systems requires large amounts of training data in the form of parallel sentences -- a resource which is often difficult and expensive to obtain, especially for languages less frequently studied in natural language processing (NLP), endangered languages, or dialects.

For some languages, it is possible to scrape data from the web \citep{resnik-smith-2003-web}, or to leverage existing translations, e.g., of movie subtitles \citep{Zhang2014DualSA} or religious texts \citep{Resnik1999TheBA}. However, such sources of data are only available for a limited number of languages, and it is impossible to collect large MT corpora for a diverse set of languages using these methods. Professional translators, which are a straightforward alternative, are often rare or expensive.

In this paper, we propose a new data collection strategy which is cheap, simple, effective and, importantly, does not require professional translators or even bilingual speakers. 
It is based on two assumptions: (1) \textbf{non-textual modalities can serve as a pivot for the annotation process} \cite{madaan2020practical}; and (2) \textbf{annotators subconsciously pay increased attention to moving objects}, since humans are extremely good at detecting motion, a crucial skill for survival \cite{albright1995visual}.
Thus, we propose to leverage graphics interchange formats (GIFs) as a pivot to collect parallel data in two or more languages.
\begin{figure}[t]
    \centering
    \includegraphics[width=1.0\columnwidth]{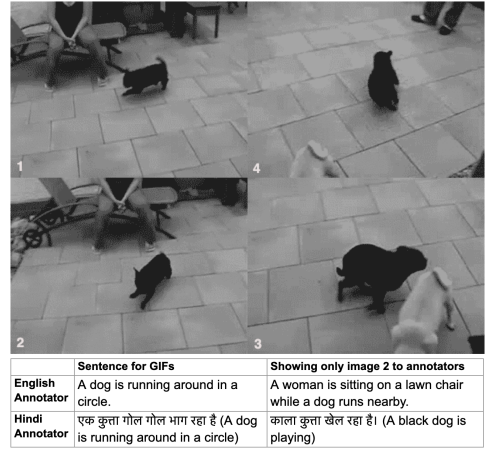}
    \caption{Sentences written by English and Hindi annotators using GIFs or images as a pivot.}
    \label{fig:dog_gif}
\end{figure}

We prefer GIFs over videos as they are short in duration, do not require audio for understanding and describe a comprehensive story visually. Furthermore, we hypothesize that GIFs are better pivots than images -- which are suggested by \citet{madaan2020practical} for MT data collection -- based on our second assumption. 
We expect that people who are looking at the same GIF tend to focus on the main action and characters within the GIF and, thus, tend to write more similar sentences. This is in contrast to using images as a pivot, where people are more likely to focus on different parts of the image and, hence, to write different sentences, cf. Figure \ref{fig:dog_gif}.

We experiment with collecting Hindi, Tamil and English sentences via Amazon Mechanical Turk (MTurk), using both GIFs and images as pivots. As an additional baseline, we compare to data collected in previous work \cite{madaan2020practical}. We perform both intrinsic and extrinsic evaluations -- by manually evaluating the collected sentences and by training MT systems on the collected data, respectively -- and find that leveraging GIFs indeed results in parallel sentences of higher quality as compared to our baselines.\footnote{All data collected for our experiments is available at \url{https://nala-cub.github.io/resources}.}

\section{Related Work}
In recent years, especially with the success of transfer learning \cite{wang-etal-2018-glue} and pretraining in NLP \cite{devlin-etal-2019-bert}, several techniques for improving neural MT for low-resource languages have been proposed \cite{sennrich2016improving, fadaee-etal-2017-data, xia2019generalized,lample2018unsupervised,lewis2019bart, liu2020multilingual}.

However, supervised methods still outperform their unsupervised and semi-supervised counterparts, which makes collecting training data for MT important.
Prior work scrapes data from the web \citep{lai2020unsupervised, resnik-smith-2003-web}, or uses movie subtitles \citep{Zhang2014DualSA}, religious texts \citep{Resnik1999TheBA}, or multilingual parliament proceedings \cite{koehn2005europarl}. However, those and similar resources are only available for a limited set of languages. A large amount of data for a diverse set of low-resource languages cannot be collected using these methods.

For low-resource languages, \citet{hasan-etal-2020-low} propose a method to convert noisy parallel documents into parallel sentences. \citet{zhang-etal-2020-parallel-corpus} filter noisy sentence pairs from MT training data.

The closest work to ours is \citet{madaan2020practical}. The authors collect (pseudo-)parallel sentences with images from the Flickr8k dataset \citep{article} as a pivot,
filtering to obtain images which are simplistic and do not contain culture-specific references. Since Flickr8k already contains 5 English captions per image, they select images whose captions are short and of high similarity to each other. Culture-specific images are manually discarded.
We compare to the data from \citet{madaan2020practical} in Section \ref{sec:evaluation}, denoting it as M20.

\section{Experiments}
\subsection{Pivot Selection}
We propose to use GIFs as a pivot to collect parallel sentences in two or more languages. As a baseline, we further collect parallel data via images as similar to our GIFs as possible. In this subsection, we describe our selection of both mediums.

\paragraph{GIFs} We take our GIFs from a dataset presented in \citet{li2016tgif}, which consists of 100k GIFs with descriptions. Out of these, 10k GIFs have three English one-sentence descriptions each, which makes them a suitable starting point for our experiments. We compute the word overlap in F1 between each possible combination of the three sentences, take the average per GIF, and choose the highest scoring 2.5k GIFs for our experiments. This criterion filters for GIFs for which all annotators focus on the same main characters and story, and it eliminates GIFs which are overly complex. 
We thus expect speakers of non-English languages to focus on similar content.

\paragraph{Images} Finding images which are comparable to our GIFs is non-trivial. While we could compare our GIFs' descriptions to image captions, we hypothesize that the similarity between the images obtained thereby and the GIFs would be too low for a clean comparison. Thus, we consider two alternatives: (1) using the \emph{first} frame of all GIFs, and (2) using the \emph{middle} frame of all GIFs. 

In a preliminary study, we obtain two Hindi one-sentence descriptions from two different annotators for both the first and the middle frame for a subset of 100 GIFs. We then compare the BLEU \citep{papineni-etal-2002-bleu} scores of all sentence pairs. We find that, on average, sentences for the middle frame have a BLEU score of 7.66 as compared to 4.58 for the first frame. Since a higher BLEU score indicates higher similarity and, thus, higher potential suitability as MT training data, we use the middle frames for the image-as-pivot condition in our final experiments.

\subsection{Data Collection}
\begin{table*}
\footnotesize
\centering
\setlength{\tabcolsep}{10pt}
\begin{tabular}{c|l}
\toprule
\textbf{Rating} & \textbf{Sentences from the GIF-as-Pivot Setting} \\
\midrule

\multirow{2}{*}{1}&A child flips on a trampoline.\\
&A girl enjoyed while playing. \\\midrule

\multirow{2}{*}{3}&A man in a hat is walking up the stairs holding a bottle of water. \\
&A man is walking with a plastic bottle. \\\midrule

\multirow{2}{*}{5}& A man is laughing while holding a gun. \\
&A man is laughing while holding a gun.
\\\midrule
\midrule
&\textbf{Sentences from the Image-as-Pivot Setting}\\\midrule
\multirow{2}{*}{1}&A woman makes a gesture in front of a group of other women. \\
&This woman is laughing. \\\midrule
\multirow{2}{*}{3}&An older woman with bright lip stick lights a cigarette in her mouth.\\
&This woman is lighting a cigarette. \\\midrule
\multirow{2}{*}{5}&A woman wearing leopard print dress and a white jacket is walking forward. \\
&A woman is walking with a leopard print dress and white coat. \\
\bottomrule
\end{tabular}
\caption{\label{ratings_sentences} Sentences obtained in English and Hindi for each setting where both annotators agree on the rating. The first sentence is the sentence written in English and the second sentence is the corresponding English translation of the Hindi sentence, translated by the authors.}
\label{tab:examplesentences}
\end{table*}
\label{sec:payment}
We use MTurk for all of our data collection. We collect the following datasets: (1) one single-sentence description in Hindi for each of our 2,500 GIFs; (2) one single-sentence description in Hindi for each of our 2,500 images, i.e., the GIFs' middle frames; (3) one single-sentence description in Tamil for each of the 2,500 GIFs; (4) one single-sentence description in Tamil for each of the 2,500 images; and (5) one single-sentence description in English for each of our 2,500 images. To build parallel data for the GIF-as-pivot condition, we randomly choose one of the available 3 English descriptions for each GIF.

For the collection of Hindi and Tamil sentences, we restrict the workers to be located in India and, for the English sentences, we restrict the workers to be located in the US. We use the instructions from \citet{li2016tgif} with minor changes for all settings, translating them for Indian workers.\footnote{Our instructions can be found in the appendix.}  

Each MTurk human intelligence task (HIT) consists of annotating five GIFs or images, and we expect each task to take a maximum of 6 minutes. We pay annotators in India \$0.12 per HIT (or \$1.2 per hour), which is above the minimum wage of \$1 per hour in the capital Delhi.\footnote{\url{https://paycheck.in/salary/minimumwages/16749-delhi}} 
Annotators in the US are paid \$1.2 per HIT (or \$12 per hour).
We have obtained IRB approval for the experiments reported in this paper (protocol \#: 20-0499).

\subsection{Test Set Collection}
\label{sec:test_set_collection}
For the extrinsic evaluation of our data collection strategy we train and test an MT system. For this, we additionally collect in-domain development and test examples for both the GIF-as-pivot and the image-as-pivot setting.

Specifically, we first collect 250 English sentences for 250 images which are the middle frames of previously unused GIFs. We then combine them with the English descriptions of 250 additional unused GIFs from \citet{li2016tgif}. For the resulting set of 500 sentences, we ask Indian MTurk workers to provide a translation into Hindi and Tamil.
We manually verify the quality of a randomly chosen subset of these sentences. Workers are paid \$1.2 per hour for this task. We use 100 sentence pairs from each setting as our development set and the remaining 300 for testing.

\section{Evaluation}
\label{sec:evaluation}
\begin{table}
\centering
\footnotesize
\setlength{\tabcolsep}{5.pt}
\begin{tabular}{c|r|r|r}
\toprule
 & \textbf{GIF-as-Pivot} & \textbf{Image-as-Pivot} & \textbf{M20} \\ \midrule
Hindi--English & \textbf{2.92} & 2.20 & 2.63 \\
Tamil--English & \textbf{3.03} & 2.33 & - \\
\bottomrule
\end{tabular}
\caption{Manual evaluation of a subset of our collected sentences; scores from 1 to 5; higher is better.}
\label{tab:results_intrinsic}
\end{table}

\begin{table}
\centering
\footnotesize
\setlength{\tabcolsep}{5.pt}
\begin{tabular}{c|c|r|r|r}
&Rating&GIF-as-pivot&Image-as-pivot&M20\\
\midrule
Hi-En&\multirow{2}{*}{5}&\textbf{13.08}&2.5&10.0\\
Ta-En&&\textbf{6.0}&3.5&-\\
\midrule
Hi-En&\multirow{2}{*}{\textgreater= 4}&\textbf{35.77}&15.5&26.43\\
Ta-En&&\textbf{37.0}&14.0&-\\
\midrule
Hi-En&\multirow{2}{*}{\textgreater= 3}&\textbf{61.15}&39.0&51.43\\
Ta-En&&\textbf{67.5}&42.5&-\\
\midrule
Hi-En&\multirow{2}{*}{\textgreater= 2}&\textbf{82.69}&63.0&75.0\\
Ta-En&&\textbf{92.5}&72.5&-\\
\midrule
Hi-En&\multirow{2}{*}{\textgreater= 1}&\textbf{100.0}&\textbf{100.0}&\textbf{100.0}\\
Ta-En&&\textbf{100.0}&\textbf{100.0}&-\\
\bottomrule

\end{tabular}
\caption{Cumulative percentages with respect to each setting; GIF-as-pivot shows the best results;}
\label{tab:results_individual}
\end{table}

\subsection{Intrinsic Evaluation}
In order to compare the quality of the parallel sentences obtained under different experimental conditions, we first perform a manual evaluation of a subset of the collected data. For each language pair, we select the same random 100 sentence pairs from the GIF-as-pivot and image-as-pivot settings. We further choose 100 random sentence pairs from M20. We randomly shuffle all sentence pairs and ask MTurk workers to evaluate the translation quality. Each sentence pair is evaluated independently by two workers, i.e., we collect two ratings for each pair. Sentence pairs are rated on a scale from 1 to 5, with 1 being the worst and 5 being the best possible score.\footnote{The definitions of each score as given to the annotators can be found in the appendix.} 

Each evaluation HIT consists of 11 sentence pairs. For quality control purposes, each HIT contains one manually selected example with a perfect (for Hindi--English) or almost perfect (for Tamil--English) translation. Annotators who do not give a rating of 5 (for Hindi--English) or a rating of at least 4 (for Tamil--English) do not pass this check. Their tasks are rejected and republished.

\paragraph{Results} The average ratings given by the annotators are shown in Table \ref{tab:results_intrinsic}.
Sentence pairs collected via GIF-as-pivot obtain an average rating of 2.92 and 3.03 for Hindi--English and Tamil--English, respectively. Sentences from the image-as-pivot setting only obtain an average rating of 2.20 and 2.33 for Hindi--English and, respectively, Tamil--English. The rating obtained for M20 (Hindi only) is 2.63. As we can see, for both language pairs the GIF-as-pivot setting is rated consistently higher than the other two settings, thus showing the effectiveness of our data collection strategy. This is in line with our hypothesis that the movement displayed in GIFs is able to guide the sentence writer's attention. 

We now explicitly investigate how many of the translations obtained via different strategies are acceptable or good translations; this corresponds to a score of 3 or higher. Table \ref{tab:results_individual} shows that 61.15\% of the examples are rated 3 or above in the GIF-as-pivot setting for Hindi as compared to 39.0\% and 51.43\% for the image-as-pivot setting and M20, respectively. For Tamil, 67.5\% of the sentences collected via GIFs are at least acceptable translations. The same is true for only 42.5\% of the sentences obtained via images.

We show example sentence pairs with their ratings from the GIF-as-pivot and image-as-pivot settings for Hindi--English in Table \ref{tab:examplesentences}.

\subsection{Extrinsic Evaluation}
\begin{table}[t]
\centering \footnotesize
\setlength{\tabcolsep}{2.5pt}
\begin{tabular}{llrrrrr}
\toprule 
\textbf{Test Set} & \textbf{Training Set} & \textbf{500} & \textbf{1000} & \textbf{1500} & \textbf{1900} & \textbf{2500}\\ \midrule
\multicolumn{7}{c}{Direction: Hindi to English}\\\midrule
GIF&GIF&\textbf{6.41}&\textbf{13.06}&\textbf{14.39}&\textbf{14.81}&\textbf{16.09}\\
GIF&Image&5.71&8.17&9.5&9.7&10.49\\
GIF&M20&3.19&6.84&7.99&6.9&N/A\\
\midrule
Image&GIF&2.93&8.18&9.11&8.84&9.24\\
Image&Image&\textbf{8.46}&\textbf{10.05}&\textbf{11.15}&\textbf{11.25}&\textbf{12.14}\\
Image&M20&1.27&5.79&6.76&6.68&N/A\\
\midrule
M20&GIF&1.66&5.21&5.75&6.78&\textbf{6.69}\\
M20&Image&1.63&4.53&4.98&5.09&5.63\\
M20&M20&\textbf{5.08}&\textbf{6.96}&\textbf{7.23}&\textbf{8.23}&N/A\\
\midrule
All&GIF&3.47&\textbf{8.46}&\textbf{9.35}&\textbf{9.81}&\textbf{10.28}\\
All&Image&\textbf{4.9}&7.28&8.19&8.32&9.04\\
All&M20&3.37&6.57&7.32&7.37&N/A\\
\midrule

\multicolumn{7}{c}{Direction: English to Hindi}\\
\midrule
GIF&GIF&0.63&1.68&2.01&1.72&\textbf{3.07}\\
GIF&Image&\textbf{0.81}&\textbf{2.18}&1.43&2.29&1.86\\
GIF&M20&0.42&2.09&\textbf{2.99}&\textbf{3.06}&N/A\\
\midrule
Image&GIF&0.11&1.19&1.03&0.97&\textbf{1.42}\\
Image&Image&0.15&1.19&1.04&1.09&1.29\\
Image&M20&\textbf{0.22}&\textbf{1.23}&\textbf{1.95}&\textbf{1.68}&N/A\\
\midrule
M20&GIF&1.15&2.75&4.25&4.52&4.88\\
M20&Image&1.32&3.09&4.41&4.1&\textbf{5.16}\\
M20&M20&\textbf{5.12}&\textbf{12.27}&\textbf{12.65}&\textbf{13.31}&N/A\\
\midrule
All&GIF&0.68&1.96&2.61&2.62&\textbf{3.3}\\
All&Image&0.82&2.25&2.51&2.65&3.01\\
All&M20&\textbf{2.24}&\textbf{5.9}&\textbf{6.54}&\textbf{6.75}&N/A\\
\bottomrule
\end{tabular}
\caption{\label{font-table} BLEU for different training and test sets; \textit{All} denotes a weighted average over all test sets; all models are obtained by finetuning mBART; best scores for each training set size and test set in bold.}
\end{table}

\begin{table}[t]
\centering \footnotesize
\setlength{\tabcolsep}{2.5pt}
\begin{tabular}{llrrrrr}
\toprule 
\textbf{Test Set} & \textbf{Training Set} & \textbf{500} & \textbf{1000} & \textbf{1500} & \textbf{2000} & \textbf{2500}\\ \midrule
\multicolumn{7}{c}{Direction: Tamil to English}\\\midrule
GIF&GIF&\textbf{2.63}&\textbf{4.46}&\textbf{8.26}&\textbf{9.27}&\textbf{4.99}\\
GIF&Image&2.33&3.34&3.00&4.77&3.83\\
\midrule
Image&GIF&0.95&2.42&3.15&3.67&2.74\\
Image&Image&\textbf{6.65}&\textbf{5.62}&\textbf{6.02}&\textbf{7.75}&\textbf{7.22}\\
\midrule
All&GIF&1.79&3.44&\textbf{5.71}&\textbf{6.47}&3.87\\
All&Image&\textbf{4.49}&\textbf{4.48}&4.51&6.26&\textbf{5.53}\\
\midrule
\multicolumn{7}{c}{Direction: English to Tamil}\\
\midrule
GIF&GIF&0&\textbf{0.54}&\textbf{1.00}&\textbf{0.83}&\textbf{0.84}\\
GIF&Image&\textbf{0.5}&0.18&0.96&0.43&0.48\\
\midrule
Image&GIF&0&0.31&0.36&\textbf{0.62}&\textbf{0.7}\\
Image&Image&\textbf{0.41}&\textbf{0.35}&\textbf{0.51}&0.36&0.29\\
\midrule
All&GIF&0&\textbf{0.43}&0.68&\textbf{0.73}&\textbf{0.77}\\
All&Image&\textbf{0.46}&0.27&\textbf{0.74}&0.4&0.39\\
\bottomrule
\end{tabular}
\caption{\label{tamil-res-table} BLEU for different training and test sets; \textit{All} denotes a weighted average over all test sets; all models are obtained by finetuning mBART; best scores for each training set size and test set in bold.}
\end{table}

We further extrinsically evaluate our data by training an MT model on it. Since, for reasons of practicality, we collect only 2,500 examples, we leverage a pretrained model instead of training from scratch. Specifically, we finetune an mBART model \citep{liu2020multilingual} on increasing amounts of data from all setting in both directions.
mBART is a transformer-based sequence-to-sequence model which is pretrained on 25 monolingual raw text corpora.
We finetune it with a learning rate of 3e-5 and a dropout of 0.3 for up to 100 epochs with a patience of 15. 
\paragraph{Results} The BLEU scores for all settings are shown in Tables \ref{font-table} and \ref{tamil-res-table} for Hindi--English and Tamil--English, respectively. We observe that increasing the dataset size mostly increases the performance for all data collection settings, which indicates that the obtained data is useful for training. Further, we  observe that each model performs best on its own in-domain test set.

Looking at Hindi-to-English translation, we see that, on average, models trained on sentences collected via GIFs outperform sentences from images or M20 for all training set sizes, except for the 500-examples setting, where image-as-pivot is best. However, 
results are mixed for Tamil-to-English translation.

Considering English-to-Hindi translation, models trained on M20 data outperform models trained on sentences collected via GIFs or our images in nearly all settings. However, since the BLEU scores are low, we manually inspect the obtained outputs. We find that the translations into Hindi are poor and differences in BLEU scores are often due to shared individual words, even though the overall meaning of the translation is incorrect. Similarly, for English-to-Tamil translation, all BLEU scores are below or equal to 1. We thus conclude that 2,500 examples are not enough to train an MT system for these directions, and, while we report all results here for completeness, we believe that the intrinsic evaluation paints a more complete picture.\footnote{We also manually inspect the translations into English: in contrast to the Hindi translations, most sentences at least partially convey the same meaning as the reference.}
We leave a scaling of our extrinsic evaluation to future work.

\section{Conclusion}

In this work, we made two assumptions: (1) that a non-textual modality can serve as a pivot for MT data collection, and (2) that humans tend to focus on moving objects. Based on this, we proposed to collect parallel sentences for MT using GIFs as pivots,  eliminating the need for bilingual speakers and reducing annotation costs. We collected parallel sentences in English, Hindi and Tamil using our approach and conducted intrinsic and extrinsic evaluations of the obtained data, comparing our strategy to two baseline approaches which used images as pivots. According to the intrinsic evaluation, our approach resulted in parallel sentences of higher quality than either baseline.

\section*{Acknowledgments}
We would like to thank the anonymous reviewers, whose feedback helped us improve this paper. We are also grateful to Aman Madaan, the first author of the M20 paper, for providing the data splits and insights from his work. Finally, we thank the members of CU Boulder's NALA Group for their feedback on this research.
\bibliographystyle{acl_natbib}
\bibliography{acl2021}

\appendix
\clearpage

\begin{table*}[t]
\section{Sentence Rating Instructions}
\centering
\footnotesize
\begin{tabular}{lll}
\hline \textbf{Score}&\textbf{Title} & \textbf{Description}\\ \hline
1&Not a translation& There is no relation whatsoever between the source and the target sentence \\
2&Bad& Some word overlap, but the meaning isn’t the same \\
3&Acceptable& The translation conveys the meaning to some degree but is a bad translation \\
4&Good& The translation is missing a few words but conveys most of the meaning adequately \\
5&Perfect& The translation is perfect or close to perfect \\
\hline
\end{tabular}
\caption{Description of the ratings for the manual evaluation of translations.}
\label{tab:rating_description}
\end{table*}

\begin{figure*}[t]
    \section{MTurk Instructions}
    \centering
    \includegraphics[width=1.0\textwidth]{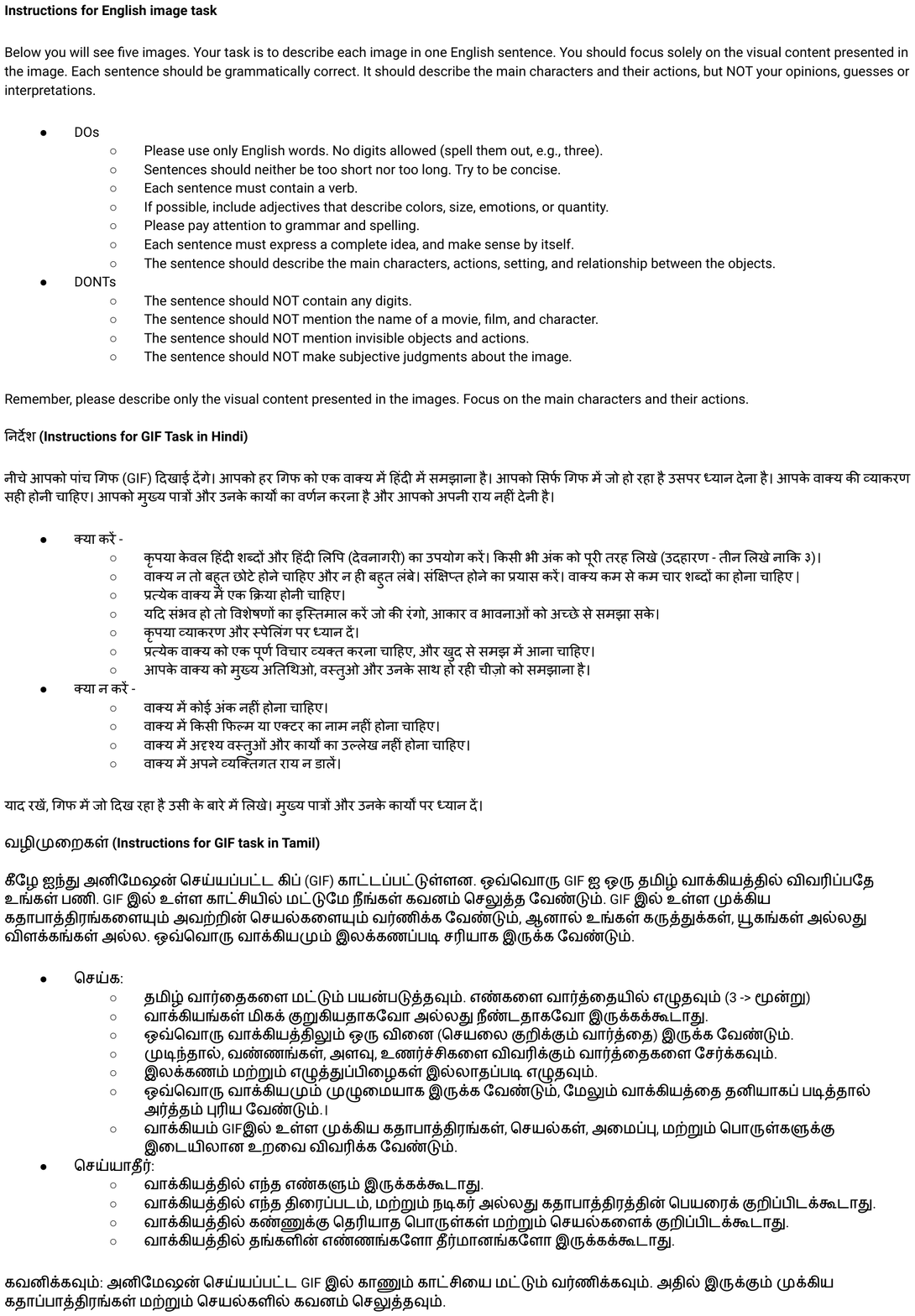}
    \caption{Instructions for the data collection via images in English, via GIFs in Hindi and Tamil.}
    \label{fig:image_task}
\end{figure*}
\end{document}